\documentclass{article}
\usepackage[margin=1in]{geometry}

\usepackage{algorithm}
\usepackage{algorithmicx}
\usepackage{algpseudocode}

\usepackage{graphicx}
\usepackage{booktabs}
\usepackage{amsmath}
\usepackage{amssymb}
\usepackage{amsfonts}
\usepackage{multirow}
\usepackage{verbatim}
\usepackage{caption}
\usepackage{longtable}
\usepackage{supertabular}
\usepackage{float}
\usepackage{paralist}

\usepackage{enumitem}
\usepackage{tablefootnote}
\usepackage[round,semicolon]{natbib}
\usepackage{xcolor}
\usepackage{xspace}
\usepackage{textcomp}
\usepackage{makecell}
\usepackage{multirow}
\usepackage{lscape} 
\usepackage{siunitx}

\usepackage{amssymb}%
\usepackage{pifont}%
\usepackage{scrextend}

\usepackage{array}
\usepackage{tgpagella}
\usepackage{latexsym}
\usepackage[T1]{fontenc}
\usepackage[utf8]{inputenc}
\usepackage{microtype}
\definecolor{mydarkblue}{rgb}{0,0.08,0.45}
\usepackage[colorlinks,citecolor=mydarkblue,urlcolor=mydarkblue,linkcolor=mydarkblue]{hyperref}
\usepackage{url}         
\usepackage{nicefrac}       
\usepackage{changepage}
\usepackage{xargs}          
\usepackage{wrapfig,lipsum,booktabs}
\usepackage{longtable}
\usepackage{endnotes}

\usepackage{pgfplots}
\usetikzlibrary{pgfplots.groupplots}
\pgfplotsset{compat=1.3}
\usepackage{tikz}
\usetikzlibrary{patterns}

\usepackage[most]{tcolorbox}

\usepackage[capitalize,noabbrev]{cleveref}
\crefname{section}{Section}{\S\S}
\Crefname{section}{Section}{\S\S}
\crefname{table}{Table}{Tables}
\crefname{figure}{Figure}{Figures}
\crefname{algorithm}{Algorithm}{}
\crefname{equation}{eq.}{}
\crefname{appendix}{Appendix}{}
\crefformat{section}{Section #2#1#3}
\usepackage{multicol}

\usepackage{minitoc}

\usepackage{microtype}
\usepackage{graphicx}
\usepackage{booktabs} %
\usepackage{hyperref}
\usepackage{siunitx}
\usepackage{amsmath}
\usepackage{amssymb}
\usepackage{mathtools}
\usepackage{amsthm}
\usepackage{fontawesome5}
\usepackage{longtable}
\usepackage{multirow}
\usepackage{subfigure}
\usepackage{caption}
\usepackage{xfrac}
\usepackage{setspace}

\usepackage{wrapfig}
\usepackage{tabulary}
\usepackage{tabularx}
\usepackage[export]{adjustbox}

\usepackage[capitalize,noabbrev]{cleveref}
\newcommand{\method}{\texttt{FedCompress}}

\title{\textbf{Communication-Efficient Federated Learning through Adaptive Weight Clustering and Server-Side Distillation}}
\author{
\normalsize{}
\textbf{Vasileios Tsouvalas$^1$ \qquad Aaqib Saeed$^1$ \qquad Tanir Ozcelebi$^1$ \qquad Nirvana Meratnia$^1$}\\
\normalsize{}
$^1$Eindhoven University of Technology, The Netherlands
}

\date{}

\begin{document}

\maketitle

\begin{abstract}
Federated Learning (FL) is a promising technique for the collaborative training of deep neural networks across multiple devices while preserving data privacy. Despite its potential benefits, FL is hindered by excessive communication costs due to repeated server-client communication during training. To address this challenge, model compression techniques, such as sparsification and weight clustering are applied, which often require modifying the underlying model aggregation schemes or involve cumbersome hyperparameter tuning, with the latter not only adjusts the model's compression rate but also limits model’s potential for continuous improvement over growing data. In this paper, we propose~\method\footnote{\url{https://github.com/FederatedML/FedCompress}}, a novel approach that combines dynamic weight clustering and server-side knowledge distillation to reduce communication costs while learning highly generalizable models. Through a comprehensive evaluation on diverse public datasets, we demonstrate the efficacy of our approach compared to baselines in terms of communication costs and inference speed.
\end{abstract}

Federated learning (FL) enables collaborative training of neural network models directly on edge devices (referred to as clients), preserving on-device data locally \cite{fl}. FL is composed of multiple federated rounds, which involve server-to-client model updates dispatch, local training by clients, and server-side aggregation (e.g., \textit{FedAvg} \cite{fedavg}) of clients' model updates, iteratively performed until model convergence. Despite its appealing properties for users' privacy, FL requires constant model transportation between server and clients, which poses a challenge in terms of communication efficiency. This becomes even more critical when the clients are resource-constrained edge devices with limited computational capabilities and strict energy constraints.

To address the communication overhead in FL, recent studies have explored model compression schemes on the exchanged model updates to minimize the communication overhead. Sparsification \cite{prunning} involves discarding network segments to reduce the overall model’s complexity based on a threshold value. FedSparsify \cite{fedsparsify} utilizes magnitude pruning with a gradually increasing threshold during training to learn a highly-sparse model in a communication-efficient FL scheme. Alternatively, weight clustering \cite{wc} converts the weight matrices elements into a discrete set of values (clusters) to achieve high model compression ratio. MUCSC \cite{mucsc} utilizes layer-wise weight clustering using a fixed number of clusters to communicate compressed model updates from clients to server. Likewise, FedZip \cite{fedzip} employs a sequence of pruning and weight clustering to further improve the compression ratio. Apart from these model compression schemes, knowledge distillation \cite{ref1} and sub-model training \cite{ref3,ref4} have being explored to reduce communication costs. However, the aforementioned techniques require modifications to the underlying model aggregation algorithm and solely focus in optimizing the client-to-server (upstream) communication route. Furthermore, existing weight clustering schemes in FL rely on a fixed set of clusters, limiting model’s potential for continuous improvement over growing data.

\begin{figure}[t]
    \centering
    \includegraphics[width=\columnwidth]{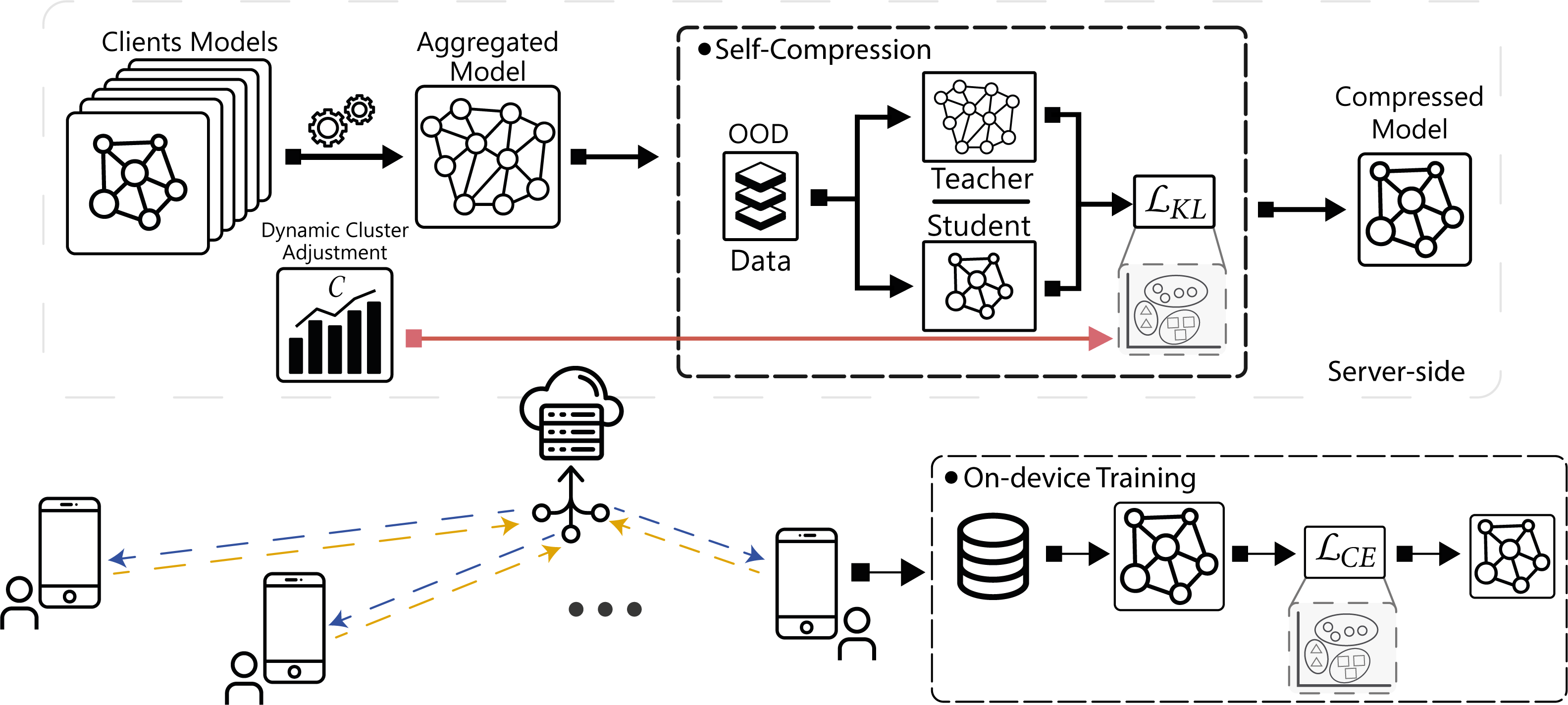}
    \caption{\footnotesize{Illustration of FedCompress for communication-efficient FL. A dual-stage compression scheme is proposed: (i) weight clustering across clients during on-device training, and (ii) self-compression on server-side combining weight clustering with knowledge distillation on out-of-distribution data.}}
    \label{fig:akd_arch}
\end{figure}

We propose~\method~(\textit{Federated Learning with Dynamic Weight Clustering for Model Compression}) to achieve significant communication reductions in the bidirectional communication route during FL training, while maintaining the representational power of models and achieving highly compressed models. Naively applying weight clustering during local updates in FL yields limited compression benefits for the aggregated model, as clients' models form diverse clusters during compression due to the statistical heterogeneity among clients. Consequently, the centroid-based structure of the aggregated model weights is compromised, once the server-side aggregation is applied \cite{briggs2020federated}. To overcome this issue with no modifications in the underlying aggregation strategy, we apply the knowledge distillation \cite{hinton_kd} scheme using out-of-distribution data after model aggregation at the server. In this way, we perform self-compression on the aggregated model, enforcing the centroid-like structure and reducing communication costs in the downstream communication channel for the next federated round. 

Aside from the server-side compression, we balance the trade-off between model performance and communication efficiency by finding a suitable number of clusters during the FL process. To this end, we start with a small number of clusters for each layer and dynamically adjust them as federated rounds progress by monitoring the representational power of clients' models. In particular, we propose a representation quality score, which is computed locally at each client using a small unlabelled set of clients' available data. Concisely, the main contributions of our work are as follows:
\begin{itemize}
\item We propose~\method, a communication efficient FL approach that combines weight clustering and knowledge distillation on out-of-distribution data to achieve highly compressed models, with no modifications to the underlying weights' aggregation algorithm.
\item We introduce a representation quality score, locally computed based on clients' unlabelled data, to dynamically adjust the number of clusters utilized during weight clustering.
\item We demonstrate that proposed method is highly effective for learning generalizable compressed federated models on diverse public datasets from both vision and audio domains, namely CIFAR-10~\cite{cifar10}, CIFAR-100~\cite{cifar10}, PathMNIST~\cite{pathmnist}, SpeechCommands~\cite{spcm} and VoxForge~\cite{voxforge}.
\item Our evaluation of~\method~shows on average a 4.5-fold reduction in communication costs during training and a 1.13-fold speedup during inference on edge accelerator devices when compared to \textit{FedAvg} across considered datasets.
\end{itemize}

\section{Methodology}\label{sec:methodology}

\subsection{Problem Formulation}
We focus on the problem of federated model compression to overcome the high communication-burden in FL and offer notable inference speedup on edge devices as a by-product. Formally, we assume that each of the $K$ clients has a labeled and a small validation (unlabeled) dataset, denoted by $\mathcal{D}_{l}^{k}$ = $\{x{i},y_{i}\}_{i=1}^{N_l^k}$, and $\mathcal{D}_{u}^{k} = \{x_{i}\}_{i=1}^{N_u^k}$, respectively. Furthermore, server possesses an OOD dataset, $\mathcal{S}= \{x_i\}_{i=1}^{N_s}$, which can have no statistical similarity with clients locally stored data. We aim to learn a compressed global model, $p_{\tilde{\theta}}$ without clients sharing locally stored data, $\mathcal{D}_{l}$ and $\mathcal{D}_{u}$, with the server and reduce the communication burden during FL training phase. Here, we denote with $p_{\theta}$ a neural network with weights $\theta$, while $\tilde{p}_{\theta}^{k}$ and $\tilde{\theta}$ refer to the weight-clustered (compressed) versions of $p_{\theta}$ and $\theta$, respectively.

\subsection{Federated Model Compression}
We propose ~\method, a two-stage compression scheme for bi-directional communication reduction during training in FL. During local model training, we simultaneously train and compress clients' models using weight-clustering to reduce the upstream communication costs. To maintain a highly compressed model once server-side model aggregation is complete, we employ a novel model compression scheme that utilizes OOD data to minimize the downstream communication burden, while maintaining high model performance. Furthermore, To strike a balance between model performance and communication efficiency trade-off, we propose a dynamically adaptive weight-clustering approach to monitor and update the number of clusters based on the representational power of clients' models, allowing~\method~to adapt to the underlying task's complexity.

\noindent\textbf{Client-side Model Compression:} During the local model update step, we initially apply standard cross-entropy to each of the $k$ client's labeled datasets, $\mathcal{D}_{l}^{k}$, while simultaneously enforcing the weights $\theta$ to cluster around a set of $C$ learnable centroids, $\mu$. Specifically, each client's minimization objective is defined as follows:

\begin{equation}\label{eqn:client_wc}
    \resizebox{0.5\textwidth}{!}{
        $
        \min_{\theta} \mathcal{L}_{k}(\theta_k) = \mathcal{L}_{ce}(p_{\theta_k}(\mathcal{D}_{l}^{k})) + \beta \cdot \mathcal{L}_{wc}(\theta_k, \mu, C)
        $
    }
\end{equation}

\noindent Here, $\mathcal{L}_{ce}$ is the cross-entropy loss function for the model $p_{\theta}$ on the labeled dataset $\mathcal{D}_{l}^{k}$. We use $\beta$ to control the relative impact of $\mathcal{L}_{ce}$ and $\mathcal{L}_{wc}$. As the initial centroids are important to maintain model's high representational power, we allow for a few training rounds using $\mathcal{L}_{ce}$ before introducing $\mathcal{L}_{wc}$. In essence, we start each local FL training step with $\beta$=$0$ for a few epochs and afterwards setting $\beta$=$1$. 

\noindent\textbf{Self-Compression on Server:} Once the server has constructed a new global model from the received model updates, the centroid-shape structure of model weights is compromised, making it challenging to maintain a compressed model for downstream communication. To solve this problem, we propose a self-compression mechanism that combines weight clustering and knowledge distillation on OOD data, $\mathcal{S}$, at the server side. This mechanism involves training a compressed version of the original global model (i.e., the teacher), which acts as a student that aims to mimic the behavior of its teacher on the OOD data. 

In this way, we enforce the model weights to cluster around a set of $C$ learnable centroids, similar to the client-side compression, while recovering any performance degradation due to weight-clustering. As a loss function to this teacher-student architecture, we utilize the Kullback-Leibler divergence (KLD) loss, which aims to match the output distributions of the models. Specifically, the objective function of the proposed server-side self-compression approach is as follows:

\begin{equation}\label{eqn:server_wc}
    \resizebox{0.62\textwidth}{!}{
        $
        \begin{aligned}
            \min_{\theta} \mathcal{L}_{s}(\theta)
                = &~\mathcal{L}_{kl} \left(p_{\theta_T}~||~p_{\theta} \right) + \beta \cdot \mathcal{L}_{wc}(\theta, \mu , C)\\
                = &~\lambda^2 \cdot \sum_{x \in \mathcal{S}} p_{\theta_T}^\lambda(x) \log \frac{p_{\theta_T}^\lambda(x)}{p_{\theta}^\lambda(x)} + \beta \cdot \sum_{j=1}^{C} \sum_{i=1}^{N_s} u_{ij} ||\theta_i - \mu_j||^2 ~~
        \end{aligned}$
    }
\end{equation}

\noindent where $\lambda$ is a temperature scalar, $\mathcal{L}_{wc}$ is the cross-entropy loss, $\mathcal{L}_{kl}$ is the KLD loss being computed using $p_{\theta_T}^\lambda$ and $p_{\theta}^\lambda$ the $\lambda$-scaled logits of teacher and student models, respectively. 

One may note that no labels are required to enforce the alignment of distributions, thus any unlabelled dataset available on the server can be utilized to perform the self-compression. Furthermore, with no modifications required to the underlying model aggregation mechanism (e.g., \textit{FedAvg}), our method provides a readily usable solution to any existing FL systems to reduce the downstream communication costs.

\noindent\textbf{Dynamic Weight-Clustering:} While model compression via weight clustering can result in highly accurate compressed models, its performance is directly affected by selecting a proper number of clusters, $C$. Especially, in FL, where clients have heterogeneous data distributions, a suitable number of clusters can vary significantly based on the heterogeneity of clients and their local data distributions. We propose dynamically adjusting the number of clusters based on clients models' representational power. In particular, we assess model's performance using embeddings from the penultimate layer of the model, where we use the rank of embeddings as a proxy of their generalization quality~\cite{rankme}. This representation quality score is computed locally on client's unlabeled dataset $\mathcal{D}_{u}^{k}$. 

Formally, with the completion of the local training step on $k$-th client, we extract the embeddings $\mathcal{Z}^k$ from $\mathcal{D}_{u}^{k}$ using client's model $p_{\theta_{k}}$, and afterwards compute the score, $\mathcal{E}_k$ using $\exp(-\sum_{j=1}^{m_{\mathcal{Z}^k}} r_{j} \log r_j )$, where $r_j$ denoting the ranking of $j$-th singular value of $\mathcal{Z}^k$ ($\frac{\sigma_j}{|\sigma_{\mathcal{Z}^k}|_1}$) and $m_{\mathcal{Z}^k}$ denotes the minimum dimension of embeddings. To ensure numerical stability, we add a small constant equal to $1e^{-7}$ in the computation of $r_j$. During the server-side model aggregation step in each federated round, we compute the weighted-averaged representation quality score $\mathcal{E}$ of participating clients models, similar to \textit{FedAvg}. 

To ensure that the model is compressed efficiently, we start with a small value for $C$ (i.e., minimum number of clusters - $C_{min}$), incrementing it when the moving average of $\mathcal{E}$ over a window $W$ shows no improvement in the previous $P$ rounds. We fix $W$=$3$ and $P$=$3$, which we determine to be working well during our initial exploration. Furthermore, our approach allows for a maximum communication budget to be specified prior to FL training (e.g. based on an energy consumption profile) to update $C$ between two boundary values, $\left [ C_{min}, C_{max} \right ]$. Further details and an overview of our proposed~\method~approach for communication-efficient FL can be found in Algorithm~\ref{alg:method}.

\section{Evaluation}\label{sec:eval}

\noindent\textbf{Datasets:} We use publicly available datasets from both the vision and audio domains with their standard training/test splits. Specifically, we use the CIFAR-10/100 \cite{cifar10} and PathMNIST \cite{pathmnist} datasets, where the tasks of interests are object detection and pathology reporting, respectively. Likewise, from the audio domain, we use SpeechCommands \cite{spcm} for keyword spotting (12 classes in total), and VoxForge \cite{voxforge} for language identification. As OOD datasets for self-compression on server, we use StyleGAN (Oriented) \cite{stylegan} and Librispeech \cite{librispeech} for our vision and audio recognition tasks, respectively. However, we note that augmented patches (or segments in case of audio) from a single image can also be used as OOD data \cite{asano2023augmented}.

\noindent\textbf{Experimental Setup:} We utilize ResNet-20~\cite{resnet} for the vision domain, while we choose MobileNet~\cite{howard2017mobilenets} for the considered audio recognition task. These models were chosen based on their performance, suitability for on-device learning, where computational resources are limited compared to centralized settings, and extensive validation in previous research~\cite{fedln}. To simulate a federated environment, we use Flower~\cite{Flower} with \textit{FedAvg}~\cite{fedavg} to construct the global model from clients' local updates. We control the federated setting in our experiments with the following parameters: \begin{inparaenum}[1)] \item  number of clients - $M$=$20$, \item number of rounds - $R$=$20$, \item local train epochs - $E_{c}$=$10$, server self-compression training epochs - $E_{s}$=$10$, \item data distribution variance across clients - $\sigma$=$25\%$. \end{inparaenum} We randomly partitioned the datasets across the available clients in a non-overlapping fashion. 

From the related approaches in the literature, we performed experiments using FedZip~\cite{fedzip} with number of clusters fixed to $15$ (which we find to work well for the considered tasks after preliminary experimentation), while compared both~\method~and FedZip performances in terms of test set accuracy, communication-cost reduction (CCR), and model compression ratio (MCR) with respect to the standard \textit{FedAvg}. For a rigorous review, we manage any randomness during data partitioning and training procedures with a seed value and performed two distinct trials, reporting the average accuracy on test sets. 

\begin{table*}[t]
    \centering
    \caption{\small{Experimental results reporting Communication Compression Reduction (CCR), Model Compression Ratio (MCR) and accuracy difference $\delta$-Acc versus standard \textit{FedAvg} for FedZip~\cite{fedzip} and~\method~(with and without Self-Compression on Server - SCS). Results are reported across five datasets, while CCR and MCR are indicating an $n$-fold reduction from the standard \textit{FedAvg} results. Federated parameters are set to $R$=20, $M$=20, $E_{c}$=$10$, $E_{s}$=$10$, and $\sigma$=25\%.}}\label{tab:results}
    \resizebox{0.75\textwidth}{!}{
        \begin{tabular}{lcccccccccc}
            \toprule
            \multirow{2}{*}{\textbf{Dataset}} & \multirow{2}{*}{\begin{tabular}[c]{c}\textbf{FedAvg}\\\textbf{Accuracy}\end{tabular}} & \multicolumn{3}{c}{\textbf{FedZip}~\cite{fedzip}} & \multicolumn{3}{c}{\textbf{FedCompress}~\textit{(w/o SCS)}} & \multicolumn{3}{c}{\textbf{FedCompress}} \\
            \cmidrule(lr){3-5} \cmidrule(lr){6-8} \cmidrule(lr){9-11}
            & & \textbf{$\delta$-Acc} & \textbf{CCR} & \textbf{MCR} & \textbf{$\delta$-Acc} & \textbf{CCR} & \textbf{MCR} & \textbf{$\delta$-Acc} & \textbf{CCR} & \textbf{MCR} \\ 
            \midrule
            \textbf{CIFAR-10} & 86.26 & -1.89 & 1.91 & 2.08 & -1.47 & 1.02 & 1.77 & -1.83 & 4.53 & 5.18 \\
            \textbf{CIFAR-100} & 60.68 & -2.57 & 1.94 & 2.11 & -2.67 & 1.02 & 1.62 & -1.88 & 3.80 & 3.93 \\
            \textbf{PathMNIST} & 88.22 & -3.04 & 1.92 & 2.10 & -3.57 & 1.06 & 1.82 & -1.72 & 4.79 & 5.27 \\
            \textbf{SpeechCommands} & 95.75 & -0.82 & 1.66 & 1.88 & -0.72 & 1.06 & 1.72 & -0.42 & 5.04 & 5.09 \\
            \textbf{VoxForge} & 81.05 & -1.04 & 1.69 & 1.91 & 0.75 & 1.11 & 1.81 & -0.31 & 5.41 & 5.64 \\
            \bottomrule
        \end{tabular}
    }
\end{table*}

\noindent\textbf{Results:} In Table~\ref{tab:results}, we report our findings across all datasets and compare~\method~performance with the considered baselines. As indicated from $\delta$-Acc (accuracy versus standard \textit{FedAvg}) columns in Table~\ref{tab:results},~\method~outperforms FedZip across all considered datasets and yields compressed federated models with comparable performance to \textit{FedAvg}, while achieving significant reductions in communication costs. On audio-based tasks, where the MobileNet model was utilized, we observe over a $5$-fold reduction in communication costs (CCR), while keeping models' accuracy within $0.5$\% of the standard FL process. In the vision domain, our approach remains equally effective, with a CCR of $4.28$, while suffering an accuracy drop of approximately $2.14$\% across all image dataset. 

\begin{figure}[!htbp]
    \centering
    \includegraphics[width=\columnwidth]{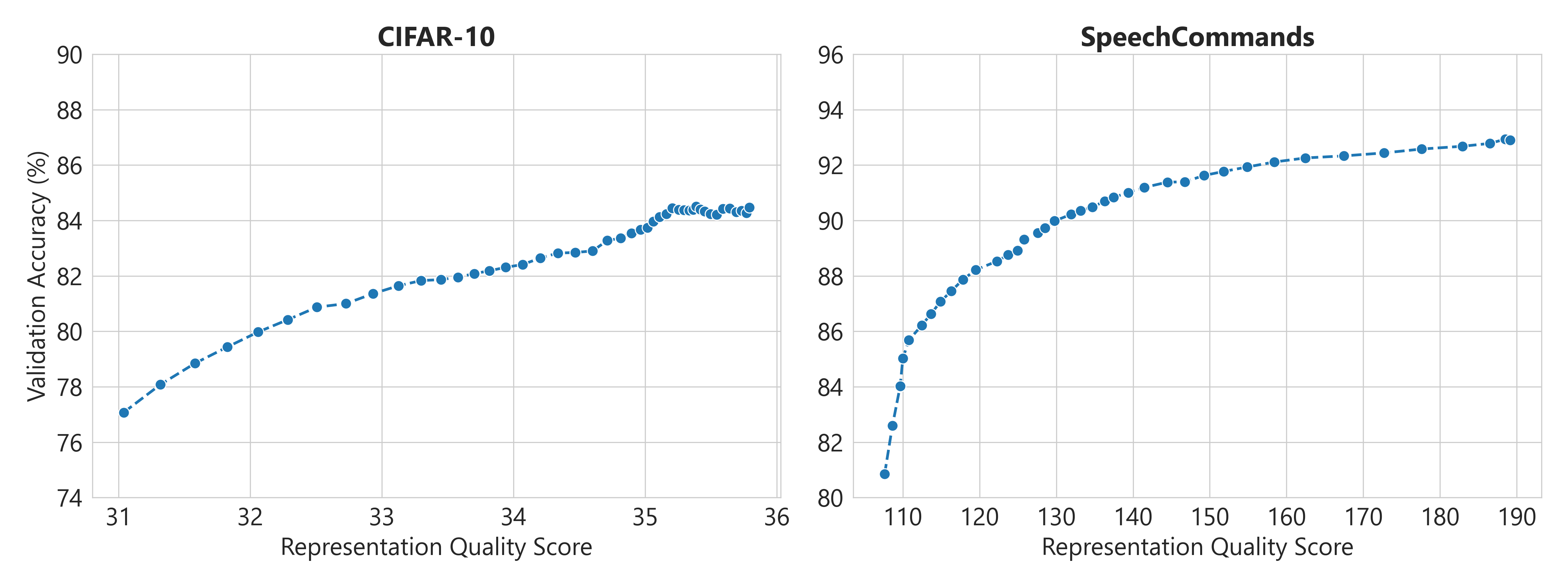}
    \caption{\small{Relationship between mean representation quality score and mean validation accuracy across clients during FL training for~\method~on CIFAR-10 and SpeechCommmands. Strong positive correlation is observed, indicating that the representation quality score is a useful indicator of the clients models' representational power.}}\label{fig:rep_score}
\end{figure}

Overall, an average $4.5$-fold CCR is achieved throughout all five datasets, suggesting that\method~can effectively reduce the communication costs in FL without sacrificing models' accuracy. These results indicate the potential for significant energy savings and reduced communication bandwidth requirements in FL, both crucial for resource-constrained devices and Internet of Things (IoT). Apart from the reduction in communication costs,~\method~demonstrates a significant reduction in resulting model size. We observe an average MCR of $4.14$ across all datasets, indicating that our approach can effectively compress models without sacrificing their accuracy. The highly compressed models offer additional benefits to edge devices, such as lower memory requirements and reduced energy consumption during training and inference. This, in turn, allows for more complex models to be deployed on edge devices with limited resources.

To validate the utilization of our representation quality score as an indicator of models' representation power, we compute the score and the validation accuracy across clients in each federated round on CIFAR-10 and SpeechCommands and compared their progression. Figure~\ref{fig:rep_score} shows a strong correlation between the two metrics, indicating that the models' representation quality metric is a valuable alternative to the validation accuracy, while it can be efficiently computed from clients models' embeddings with no need for having labeled data. Since unlabeled data are often readily available on edge devices, our score can effectively act as a proxy to dynamically set the number of clusters during training, providing a lightweight and fast alternative for measuring models' representational power.

We also conducted experiments on various edge devices to evaluate the impact of our compression approach on inference time. Table~\ref{tab:inference} shows that our approach can accelerate the inference time on all these devices, while maintaining comparable accuracy to \textit{FedAvg} as shown in Table~\ref{tab:results}. Notably,~\method~models demonstrate an acceleration of up to 1.15{\small$\times$} across the edge devices, while quantizing the models to uint8 achieves an inference speedup up to 1.24{\small$\times$}. Consequently,~\method~provides the ability to achieve faster inference times with minimal effect on federated models' accuracy, providing significant benefits for resource-constrained edge devices, where low power consumption and reduced inference time are crucial.

\begin{table}[t]
    \centering
    \caption{\small{Inference time acceleration of diverse edge devices for ResNet-20 and MobileNet on CIFAR-10 and SpeechCommands datasets, respectively. The reported inference time acceleration is achieved by comparing to FL models using \textit{FedAvg}.}}\label{tab:inference}
    \resizebox{0.42\textwidth}{!}{
        \begin{tabular}{llcc}
            \toprule
            \textbf{Model} & \textbf{Device} & \textbf{float32} & \textbf{uint8 (Quantized)} \\
            \midrule
            \multirow{2}{*}{ResNet-20} & Pixel 6    & {\small$\times$}1.103 & {\small$\times$}1.165 \\
                                     & Jetson Nano  & {\small$\times$}1.127 & {\small$\times$}1.169 \\
                                     & Coral TPU    & {\small$\times$}1.113 & {\small$\times$}1.191 \\
            \midrule
            \multirow{2}{*}{MobileNet} & Pixel 6       & {\small$\times$} 1.114 & {\small$\times$}1.248 \\
                                    & Jetson Nano   & {\small$\times$}1.137 & {\small$\times$}1.161 \\
                                    & Coral TPU     & {\small$\times$}1.152 & {\small$\times$}1.194 \\
            \bottomrule
        \end{tabular}
    }
\end{table}

\section{Conclusion}\label{sec:conclusion}
We presented~\method, a communication-efficient FL approach based on model-compression via weight clustering and knowledge distillation. It can be easily integrated with existing FL frameworks, requiring no modifications to the FL aggregation strategy. Our experiments across multiple datasets showed that our approach can achieve significant reductions in communication costs and model sizes, while maintaining comparable accuracy to the standard FL process. Moreover, we have shown that our proposed embeddings-based representation quality score can effectively act as a proxy for models' representational power, allowing for a dynamic adjustment of the number of clusters during training based on the underlying task's complexity.

\section*{Acknowledgement}
\small{This research is partially funded by the DAIS project, which has received funding from KDTJU under grant agreement No 101007273.}

\bibliographystyle{plainnat}
\bibliography{ref}

\newpage
\appendix
\onecolumn

\section{\method~Algorithm}\label{asec:algo}

We provide the pseudocode for~\method~in Algorithm~\ref{alg:method}.

\begin{algorithm}[!h]
    \centering
    \caption{\small{~\method: Federated Learning with Adaptive Weight Clustering and Server-Side Distillation for Model Compression. We develop a two-stage model compression approach to reduce the communication burden in FL and offer inference speedups in edge devices. FedAvg~\cite{fedavg} is the base algorithm, whereas $\eta_{s}$ and  $\eta_{c}$ refers to the learning rates of server and clients, respectively.\label{alg:method}}}

    \begin{algorithmic}[1]
        \State Server initialization of model with model weights $\theta^{0}$, $C$=$C_{min}$
        \For{ $i=1,\dots,R$ }
            \State Randomly select $K$ clients to participate in round $i$
            \For{ each client $k \in K$ \textbf{ in parallel}}
                \State ($\theta_{k}^{i+1}$, $\mathcal{E}_{k}^{i+1}$) $\gets$ $\textbf{ClientUpdate}$($\theta^{i}$,$C_{i}$) 
            \EndFor
            \State $\theta^{i+1} \gets \sum\nolimits_{k=1}^{K} \frac{N_k}{N}~~\theta_k^{i+1}$,~~$\mathcal{E}^{i+1} \gets \sum\nolimits_{k=1}^{K} \frac{N_k}{N}~~\mathcal{E}_k^{i+1}$
            \State $\theta^{i+1}$ $\gets$ $\textbf{SelfCompress}$($\theta^{i}$,$C_i$)            
            \State $C_{i+1} \gets C_{i}  + \operatorname{sgn}\left|\operatorname{MA}(\mathcal{E}^{i+1}) - \min \limits_{j=1}^{P}\operatorname{MA}(\mathcal{E}^{i-j+1})\right|$
        \EndFor
        \Procedure{ClientUpdate}{$\theta$,$C$}
            \For{epoch $e=1,2,\dots,E_{c}$}
                \For{ batch $b \in \mathcal{D}_{l}$}
                    \State $\theta \gets \theta - \eta_{c} \cdot \nabla_{\theta} \left( \mathcal{L}_{ce} \left( p_{\theta}\left(b\right)\right) + \beta \cdot \mathcal{L}_{wc}\left(\theta, \mu, C \right) \right)$
                \EndFor
            \EndFor
            \State $\mathcal{E} \gets \mathcal{E} (p_\theta(\mathcal{D}_{u}))$
        \State \Return ($\theta$, $\mathcal{E}$)
        \EndProcedure
        \Procedure{SelfCompress}{$\theta$,$C$}
            \For{epoch $e=1,2,\dots,E_{s}$}
                \State $\theta^{\star} \gets \theta$
                \For{ batch $b \in \mathcal{S}$}
                    \State $\theta \gets \theta - \eta_{s} \cdot \nabla_{\theta} \left( \mathcal{L}_{kl} \left(p_{\theta^{\star}}(b)~||~p_{\theta}(b) \right) + \beta_{s} \cdot \mathcal{L}_{wc}(\theta, \mu, C) \right)$
                \EndFor
            \EndFor
            \State \Return $\theta$
        \EndProcedure
    \end{algorithmic}
\end{algorithm}

\end{document}